\documentclass[final,16pt]{elsarticle}

\usepackage{graphicx}
\usepackage[table]{xcolor}
\usepackage{pgfplots}
\usepackage{tikz}
\usepackage{amssymb}
\usepackage{amsmath}
\usepackage{lipsum}
\makeatletter
\def\ps@pprintTitle{%
 \let\@oddhead\@empty
 \let\@evenhead\@empty
 \def\@oddfoot{}%
 \let\@evenfoot\@oddfoot}
\makeatother

\begin{document}

\begin{frontmatter}


\title{\textbf{Photo-Quality Evaluation based on Computational Aesthetics: Review of Feature Extraction Techniques
}}


\cortext[cor1]{Semester project literature review for the MSc course Computer Vision by Ioannis Pitas.}



\author{\textbf{Dimitris Spathis}\corref{cor1}}
\ead{sdimitris@csd.auth.gr}
\address{Department of Informatics, Aristotle University of Thessaloniki, Greece}

\begin{abstract}
Researchers try to model the aesthetic quality of photographs into low and high-level features, drawing inspiration from art theory, psychology and marketing. We attempt to describe every feature extraction measure employed in the above process. The contribution of this literature review is the taxonomy of each feature by its implementation complexity, considering real-world applications and integration in mobile apps and digital cameras. Also, we discuss the machine learning results along with some unexplored research areas as future work.

\end{abstract}


\end{frontmatter}


\section{Introduction}
\label{S:1}

Quality assessment of photographs is not a new issue arising with digital cameras. Birkhoff, back in 1933 proposed that the aesthetic appeal of objects relates to the ratio of order and complexity in images. The hardness in this task is to define order and complexity [4]. With the advent of digital cameras and smartphones, people capture more photographs than they can consume. Nowadays, social media provide an adequate filter of our social graph, displaying to us photographs and posts by our friends or friends of friends, based on popularity metrics. What if we could see photographs based on our taste, previous likes, or general aesthetic criteria? In the field of information retrieval, the content based image retrieval (CBIR) systems apply computer vision techniques to the image retrieval problem, by searching in large databases. In this context, we do not search for metadata or documents, but we decompose the image on its pixels, attempting to draw inferences between its content and its aesthetic value. This relatively recent research area is termed as “Computational Aesthetics”. Researchers in this area leverage techniques from image processing and computer vision, combining methodologies from psychology and art theory. [32,22]

The field of Computational Aesthetics gradually attracts interest in the scientific community. Scopus data (figure 1) reveal the steady growth in literature regarding this area. In 2014, 89 papers were published in international conferences and journals. If we dig more on Scopus data, we find that the majority of publications comes from Asia (National University of Singapore, Universiti Tenaga Nasional and Zhejiang University) and North America (Simon Fraser University, Carnegie Mellon University, and Georgia Institute of Technology).

\begin{center}
\begin{tikzpicture}

	\begin{axis}[
    	width=15cm,
        height=5cm,
    	xtick=data,
        xticklabel style=
{/pgf/number format/1000 sep=,rotate=60,anchor=east, font=\scriptsize},
		xlabel=,
		ylabel=Papers on Scopus]
	\addplot[color=red,mark=x] coordinates {
		(1990, 1)
        (1991, 2)
        (1992, 1)
        (1993, 2)
        (1994, 2)
        (1995, 5)
        (1996, 13)
        (1997, 7)
        (1998, 9)
        (1999, 8)
        (2000, 7)
        (2001, 11)
        (2002, 27)
        (2003, 19)
        (2004, 16)
        (2005, 29)
        (2006, 34)
        (2007, 33)
        (2008, 43)
        (2009, 37)
        (2010, 56)
        (2011, 45)
        (2012, 71)
        (2013, 74)
        (2014, 89)
        (2015, 67)
		
	};
	\end{axis}
    
\end{tikzpicture}
\begin{figure}[h]
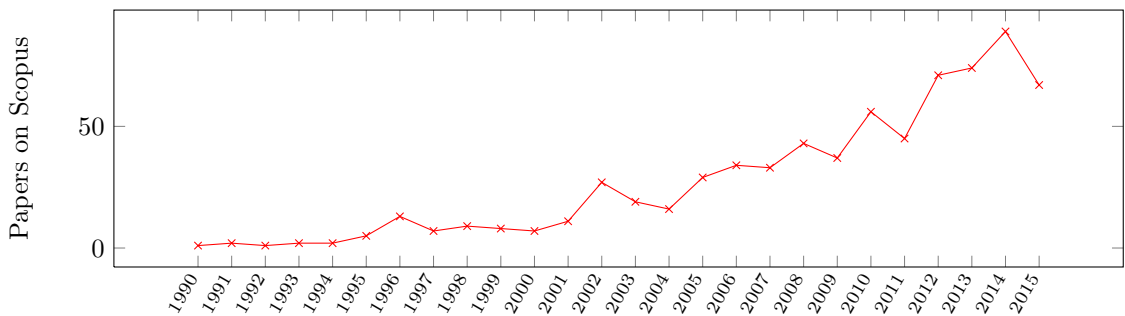

\setlength{\abovecaptionskip}{-10pt}
\caption{Papers on Scopus containing "Computational Aesthetic[s]" in abstract, title, or keywords}
\end{figure}
\end{center}

Researchers involved in aesthetic assessment of photographs try to correlate low or high-level attributes of an image with its quality evaluation. This process, understandably contains subjectivity in gathering ground truth data. The solution proposed is to combine these extracted features with crowdsourced ratings of photographs, ideally of big populations [30].

But, before we can design features to assess quality, we should decide on the perceptual criteria that we use when we judge photographs. Literature groups three factors on judging the quality of photographs: simplicity, realism and basic photography techniques [15].

\textbf{Simplicity}. One distinctive aspect of professional shots to everyday snapshots is that, pro photos are simple and easy to separate the foreground from the background. The tricks used by professionals in order to distinguish the background from the subject is by widening the lens aperture, increasing the natural color contrast of the subject and the lighting contrast.

\textbf{Realism}. Our snapshots usually look ordinary and quite common, while the professional shots look surreal. To achieve this effect, professionals are aware of the lighting conditions of dusk or dawn, choosing to shoot at hours that the sunlights are indirect. Also, they use hardware filters to make the sky bluer (ND filters), adjusting the color palette and saturation levels. Hardware in general, is more expensive and customized than regular smartphones, allowing to tweak the aperture, shutter speed, lens optical zoom and lots of other options. By leveraging the above techniques, professionals tend to shoot more unusual objects and situations.

\textbf{Basic photography techniques}. It is very rare for a professional photo to be entirely blurry. Blurriness is often the result of camera shake or low quality lens. Background blurriness, though, could be an aspect of professional deliberate result, as described above. In addition, professionals shoot usually higher contrast photos than point-and-shoot snapshot users.

Given the above grouping, researchers try to model these factors into features that can be extracted from images. We will describe those features in section 2. The rest of the paper is structured as follows: in section 3 we describe the aesthetic benchmark datasets used in computational evaluation, in section 4 we asses the machine learning frameworks of predicting aesthetic value and, finally, in section 5 we discuss the limitations and future work.

\section{Feature Extraction }
\label{S:2}

In the last fifteen years there have been significant contributions to the field of image representation and feature extraction towards semantic understanding. [14] Aesthetics and emotional value are based on semantics, so it is not surprising that they draw inspiration from the above fields. In the literature we notice a spectrum of low level (edges, textures, color histograms etc.) and high level (rule of thirds, symmetry, saliency, face recognition etc.) features combined. We group all the available features mentioned in literature as follows: \textbf{\textit{color, texture, composition, content}}. Also, we attempt a qualitative labelling of each feature according to its implementation complexity with current image processing tools and frameworks. Note that the implementation complexity might overlap with computation complexity particularly in data intensive extraction techniques, such as facial recognition. 

\subsection{Color}
\textbf{HSL and HSV color spaces}. Hue, saturation and lightness (HSL) and hue, saturation and value (HSV) are the two most common cylindrical-coordinate representations of points in an RGB color model. They are used in literature due to the better representation of color than RGB. Features often extract the average value of each channel in the whole image, or segments of it.

\textbf{White Balance.} Professional cameras have to take into account the “color temperature” of the light source. On the other hand, cheap cameras tend to shoot blue-ish photographs due to low-quality lenses. We estimate the average color temperature distribution.

\textbf{Contrast}. One of the most important aspects of a photograph is the difference in colour luminance that makes an object or thing distinguishable. We create a multi-scale contrast map from the brightness histogram of the photo and count average and median values.

\textbf{Pixel Intensity}. Too much exposure often creates lower quality pictures. Those that are too dark are often also not appealing. Thus light exposure can often be a good discriminant between high and low quality photographs. However, an over/under-exposed photograph under certain scenarios may yield very original and beautiful shots. [8] We use the average pixel intensity to estimate the use of light, where Iv is the Value channel of HSV from each row X, column Y of the image.

\begin{equation}
f=\frac{1}{XY}\sum_{x=0}^{X-1} \sum_{y=0}^{Y-1} I_{V}(x,y)
\end{equation}

\textbf{Pleasure, arousal, dominance}. Emotional coordinates based on saturation and brightness are estimated by [32] from their psychology experiments. This relationship between the saturation and brightness is modelled as follows: 

\begin{equation}
Pleasure = 0.69 V + 0.22 S
\end{equation}
\begin{equation}
Arousal = 0.31 V + 0.60 S
\end{equation}
\begin{equation}
Dominance = -0.76 V + 0.32 S
\end{equation}
where V, S are the Value and Saturation metrics of the HSV color model.

\textbf{Color Templates}. There are several metrics proposed by [18] in order to estimate the distance between the hue distribution and a certain color template. In particular, Munsel color system provides efficient regions in a color wheel according to its position. In (fig 2) we see the hue distribution models, where the gray color indicates the efficient regions that result in harmony.

\includegraphics[width=\textwidth]{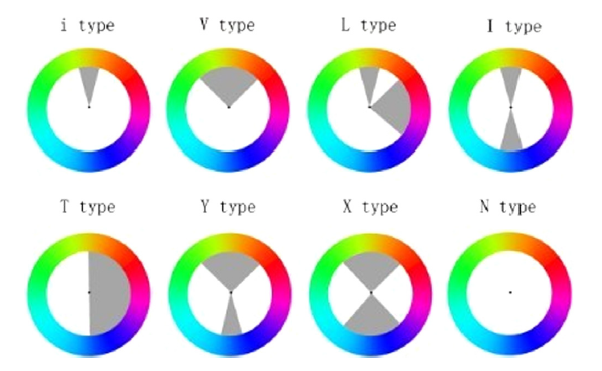}
\begin{figure}[h]
\setlength{\abovecaptionskip}{-10pt}
\caption{Munsel color wheel templates. Gray areas are aesthetically efficient.}
\end{figure}

[35] proposed an improved metric by approximating the color distribution of hue histogram by identifying its location of peak value. The distance between the hue histogram distribution and the most matching Munsel color template is expressed as: 

\begin{equation}
D_{k}=\sum_{0}^{255} dist(i,R_{k}) S(i)
\end{equation}

\begin{equation}
 dist(H(i), R_{k} =
    \begin{cases}
      0, & \text{ if i in Rk}\\
      H(i) arc(I,R_{k}) 	, & \text{otherwise}
    \end{cases}
\end{equation}
where 	k=1...7 is the type of color template (see above fig 2), dist(i,Rk) is the distance between hue value i and the gray region Rk of template Tk, H(i) stands for the occurrence of hue value i in the histogram, dist(H(i), Rk) is zero when i is in the gray regions Rk arc=length distance between i and the nearest border of gray regions and S(i) is the average saturation of hue value i  and acts as weight on colors with low saturation.

Then, we should find which template fits better our test image. Ideally this is the color template with the minimum distance. However some templates contain more gray than others, so there must be some bias. We overcome this by measuring the fitness of color templates

\begin{equation}
R_{k}= n_{k} / p_{k} ||I_{s}||
\end{equation}
where	nk is the total number of pixels in gray regions, Is the image size and pk the inverse proportion of gray region size in the template. Combining the ratio Rk and distance Dk, we measure the similarity between hue histogram and color templates: 
\begin{equation}
S_{k} = D_{k} \exp(-\sigma R_{k})
\end{equation}

\textbf{Color moment}. A group of standard measures which characterise color distribution in an image in the same way that central moments describe a probability distribution uniquely. We measure mean, standard deviation, skewness and kurtosis, for each HSV channel. 

\textbf{Colourfulness}. [8] proposed a robust method to find the relative color distribution, distinguishing colorful images from monochromatic and low contrast ones. They employed the Earth Mover’s Distance (EMD) which is a measure of similarity between any two weighted distributions. Dividing the colour space into n cubic blocks with four equal partitions along each dimension, they take each such cube as a sample point. Distribution D1 is generated as the color distribution of a hypothetical image such that for each of n sample points, the frequency is 1/n. Distribution D2 is computed from the test image, by finding the frequency of occurrence of color within each of the n cubes. But, EMD requires the pairwise distance between the two under-comparison points, so we count the pairwise Euclidean distances between the geometric centers C of each cube. Finally the colourfulness measure is estimated as follows: 

\begin{equation}
f = emd (D_{1}, D_{2} (d(a,b)\quad | \quad  0 \leq a,b \leq n-1)) \quad \text{where} \quad d(a,b)= ||Ca-Cb||
\end{equation}

\textbf{Colour Names}. We consider that every pixel of an image can be assigned to one of the following groups: black, blue, brown, green, gray, orange, pink, purple, red, white, yellow. The algorithm proposed by [33] mimics the way humans judge a whole photo of an image by its color. This measure let us decide on the style of the photographer as well.

\begin{center}
\begin{table}[h]
\begin{tabular}{l l l l}
\hline
\textbf{Feature} & \textbf{Description} & \textbf{Proposed by} & \textbf{Implementation}\\
\hline
HSL, HSV &hue, saturation, lightness & &\cellcolor{green!25}easy   \\
White Balance &color temperature & &\cellcolor{green!25}easy   \\
Contrast & multi-scale contrast map & &\cellcolor{green!25}easy   \\
Pixel Intensity & use of light of  the V channel & Datta et al &\cellcolor{green!25}easy   \\
Pleasure/Ar/Dom& emotional coordinates & Valdez et al
&\cellcolor{green!25}easy   \\
Color Templates &Munsell templates distance&Li  et al &\cellcolor{orange!25}medium   \\
Color Moment & color statistics & Datta et al  &\cellcolor{green!25}easy   \\
Colourfulness & color diversity dstance & Datta et al &\cellcolor{orange!25}medium   \\

Colour Names  & amount of specific colors & van de Weijer et al &\cellcolor{green!25}easy   \\
Bags-of-color & color harmony in local regions & Nishiyama et al &\cellcolor{red!25}hard   \\
Itten Contrasts & warm, cold contrasts & Itten &\cellcolor{orange!25}medium   \\
Dark Channel & detect haze area & He et al. &\cellcolor{orange!25}medium   \\
\hline
\end{tabular}
\caption{Color Features}
\end{table}
\end{center}

\textbf{Bags-of-color}. [23] proposed that a photograph can be seen as a collection of local regions with color variations that are relatively simple. Drawing inspiration from bags-of-words models used in NLP, they coined their term as bags-of-color-patterns. In particular, the proposed method:
\begin{enumerate}
\item samples local regions of a photo using grid sampling, distinguishing uniform regions from regions around edges and corners. Segmentation is done by mean shift, and the color boundaries detection by discriminant analysis.
\item describes each local region by features based on color harmony in Munsel color space. 
\item quantizes these features by using visual words in codebooks, by using k-means clustering.
\item represents the photo as a histogram of quantized features
\end{enumerate}

\textbf{Itten Contrasts}. A concept from art theory. [13] studied the usage of color in art and he formalized contrast concepts as to combine colors that trigger an emotional effect in the observer. Itten proposed a spherical harmony model which can be interpreted in many ways in order to extract harmonic colors, as opposing colors in rectangle or triangle. The features we measure is average contrast of brightness, contrast of saturation, contrast of hue, contrast of complements, contrast of warmth, harmony, hue count, hue spread, area of warm, area of cold, and the maximum of each. [22] provides a detailed estimation process.

\textbf{Dark Channel}. Haze effect was described by [11] as the dark channel method which identifies the ill-focused  or dull color layout in many amateur photos that suffer from an effect resembling a cloud of haze of the image. They proposed that at least one color channel has more than one pixel close to zero in the haze-free area. [35] improved this feature by taking into account the reduced Depth of Field of professionals and the foreground - background change.

\subsection{Texture}
\textbf{Tamura}. A set of various features proposed by [31] based on psychological experiments. They are successfully used in affective image retrieval. The measure computes distances of notable spatial variations of grey levels. The most common Tamura features used in literature are coarseness, contrast and directionality. 

\textbf{Edges}. Edge detection is one of the most common image processing tasks for semantic understanding. In our literature it is widely accepted and implemented in multiple forms, mainly in CANNY and SOBEL filters. In some cases [25], where it was used for architecture images aesthetics, a line-specific edge detection filter is employed in order to estimate the optimal angles of buildings. They measured vertical, horizontal and nondirectional edges. Generally, we expect the edges in professional photos to be clustered near the center of the image.

\begin{center}
\begin{table}[h]
\begin{tabular}{l l l l}
\hline
\textbf{Feature} & \textbf{Description} & \textbf{Proposed by}& \textbf{Implementation}\\
\hline
Tamura &coarseness contrast directionality &Tamura et al &\cellcolor{green!25}easy   \\
Edges &CANNY / SOBEL  & &\cellcolor{green!25}easy   \\
Spatial Envelope  & Gabor filters &Oliva et al &\cellcolor{red!25}hard   \\
Wavelet Blurriness & wavelet textures  & Datta et al  &\cellcolor{orange!25}medium   \\
GLCO Matrix& contr/corr/enrgy/hmg  & Machajdik et al
&\cellcolor{green!25}easy   \\
Visual words &interest points & Bay et al &\cellcolor{red!25}hard   \\
\hline
\end{tabular}
\caption{Texture Features}
\end{table}
\end{center}

\textbf{Spatial Envelope}. [24] proposed the GIST measure as a low-level scene descriptor. A set of dimensions, that represent the high-level structure of a scene (naturalness, openness, roughness, expansion, ruggedness) is estimated using spectral information and coarse localization. In particular, the image is segmented in a 4x4 grid and a histogram of gradients is computed for each region, and color channel.

\textbf{Wavelet Blurriness}. [8] proposed that the use of texture is a skill in photography.  By measuring the Daubechies wavelet transform they estimate the spatial smoothness. They performed a 3-level wavelet transform on all three color channels of HSV.

\textbf{Gray-Level Co-Occurrence Matrix}. A classic method for measuring texture changes. We measure contrast, correlation, energy and homogeneity for Hue, Saturation and Brightness channel.

\textbf{Visual words}. In order to capture spatial information of an image, we use a popular approach for exploiting local edges in images that creates visual words. The local edges correspond to change in intensity and are considered as interest points. By following the codebook approach as in Bags-of-color, the image can be considered as a bag of visual words. In literature [25], the Speeded Up Robust Features (SURF) framework [2] is used, which is based on Haar wavelets.

\subsection{Composition}

\textbf{Level of Detail}. Images with detail generally trigger different emotional effects than minimal shots. To measure this feature,[22] counted the number of regions after a waterfall segmentation. That way, we can distinguish between cluttered images and more simpler ones.

\textbf{Blur}. [15] modelled the blurriness of a photo as the result of a Gaussian smoothing filter applied to an otherwise sharp image, The challenge is to recover the smoothing parameter given only the blurred image. The image quality would be inversely proportional to this parameter. We can estimate the maximum frequency of the blurred image by taking its 2D Fourier transform and
count the number of frequencies whose power is greater than some threshold. The final quality of our test image is estimated by the ratio of the Fourier set of frequencies by the image size.

\textbf{Dynamic Lines}. [13] noted that lines in images induce emotional triggers. Horizons are depicted by horizontal lines and communicate calmness and relaxation, while vertical lines are clear and straightforward. Leaning lines, communicate dynamism or uncertainty. Line length and thickness overstate the above effects. By using the Hough transform we can detect those patterns and classify them as static or slant. 

\textbf{Shape Convexity}. Humans tend to display different emotions when are shown rounded or convex objects. [8] We segment our image in patches and compute their convex hull. For a segment to be considered a perfect convex shape, its segment by convex hull ratio must be 1. The segmentation process is the most critical part of the success of this algorithm. The final feature is the fraction of the image covered by approximately convex-shaped homogeneous regions

\textbf{Rule of Thirds}. Photographers trying to emulate the golden ratio in their shots, tend to shoot their subjects at one of the four intersections of the inner rectangle of the viewfinder. This implies that a large part of the interest point often lies on the periphery or inside the inner rectangle. [8] modelled this feature as:

\begin{equation}
f=\frac{9}{XY}\sum_{x=X/3}^{2X/3} \sum_{y=Y/3}^{2Y/3} I_{h}(x,y)
\end{equation}
with Ih being the Hue value of the HSV color mode. Respectively, we measure the S and V.

\textbf{Uniqueness and Familiarity}. We humans learn to judge the aesthetics of pictures from the experience gathered by seeing other pictures. Our opinions are often predefined by what we have
seen in the past. When we see something rare we perceive it very different than normal situations. We model this uniqueness feature as the integrated region matching (IRM) image distance [18] The IRM distance computes image similarity by using color, texture and shape information from automatically segmented regions, and performing a robust region-based matching with other images. According to our datasets we can compare unseen test photos with our aesthetic-extracted ground truth, in order to decide if it is unique compared to them. 

\textbf{Size and Aspect Ratio}. These simple features measure the width and height of our image. The aspect ratio especially, is important for optimal aesthetic cropping [3,5,17] 

\begin{center}
\begin{table}[h]
\begin{tabular}{l l l l}
\hline
\textbf{Feature} & \textbf{Description} & \textbf{Proposed by}& \textbf{Implementation}\\
\hline
Level of Detail &waterfall segmentation &Machajdik et al
 &\cellcolor{green!25}easy   \\
Blur  &magnitude-based frequency &Ke et al
 &\cellcolor{red!25}hard   \\
Dynamic Lines &static dynamic thick lines  &Itten &\cellcolor{green!25}easy   \\
Shape Convexity &object roundness &Datta et al &\cellcolor{orange!25}medium   \\
Rule of Thirds &inner rectangle &Datta et al &\cellcolor{green!25}easy   \\
Uniqueness &integrated region matching &
Li et al
 &\cellcolor{red!25}hard   \\
Size, Aspect Ratio&width, height & &\cellcolor{green!25}easy   \\

Image Complexity &Shannon Entropy &Romero et al
 &\cellcolor{green!25}easy   \\
 Processing Complexity &
Kolmogorov complx., Zurek entr.
 &Machado et al &\cellcolor{orange!25}medium   \\
 Visual weight ratio &geometric context &Hoiem et al.
 &\cellcolor{green!25}easy   \\
 Foreground position &foreground and rule of 3ds &Bhattacharya et al
 &\cellcolor{green!25}easy   \\
 
 Salient regions &interest regions &
Wong et al &\cellcolor{red!25}hard   \\
 Graphlets &regions as graphs  &
Zhang et al

 &\cellcolor{red!25}hard   \\
 Pyramid of HOG &self-similarity complexity anisotropy &
Redies et al
 &\cellcolor{orange!25}medium   \\
 
Symmetry&HOG  &
Dalal et al
&\cellcolor{green!25}easy   \\

\hline
\end{tabular}
\caption{ Composition Features}
\end{table}
\end{center}

\textbf{Image and Processing Complexity}. These features are inspired by Birkhoff’s idea that aesthetic appeal of objects relates to the ratio of order and complexity. The optimal image is accomplished by an image that has a low image complexity and a low processing complexity, at the same time. [27,28,21] The idea states that compression error correlates with complexity. They employ Shannon's entropy in order to measure the Image complexity. The Processing complexity is estimated via the Kolmogorov complexity.

\textbf{Visual Weight Ratio}. [12] proposed a method to recover the surface layout from an outdoor image using geometric context. The scene is segmented into sky regions, ground regions, and vertical standing objects, using adaboost on a variety of low level image features. We take vertical standing objects as subject areas. That way we distinguish between clear, cloudy and sunset skies. The ratios between the areas of these foreground-background rectangles should be close to the golden ratio for a better appeal. 

\textbf{Foreground Position}. [3] modelled the relative foreground position as the normalized Euclidean distance between the foreground’s center of mass (visual attention center), to each of four symmetric stress points (rule of thirds intersections)  in the image frame. This metric is ideal for single-subject scenes such as animals or portraits but might me ineffective in landscapes or seaside view which do not have a clear foreground. For these situations we choose the above technique of Visual weight ratio.

\textbf{Salient regions}. Saliency of a subject is the quality by which it stands out compared to its neighbors. [37] extract the salient regions from an image by utilizing a visual saliency model. We assume that the salient regions contain the photo subject. We first find  the salient locations in the original image, compute the saliency map, the segmented image, and end up with the salient mask based on the main salient locations. Our target is to enhance those salient regions as foreground objects. [38]  This saliency retargeting problem (Fig 3) is solved by Sequential Quadratic Programming (SQP).
\begin{center}
\includegraphics[width=15cm]{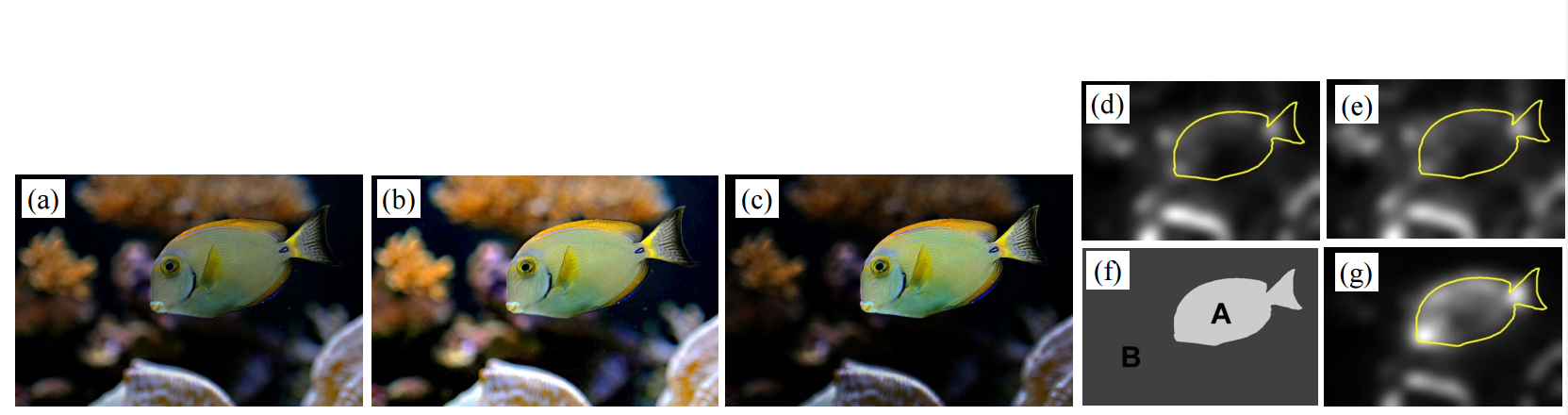}
\end{center}
\begin{figure}[h]
\setlength{\abovecaptionskip}{-10pt}
\caption{\textbf{Saliency retargeting process}: (a), (d) Original image and its saliency map. (b), (e) Globally-enhanced image and its saliency map. (c), (g) Image enhanced by saliency retargeting and its saliency map. (f) Object segments, where Objects A and B are in decreasing order of importance. Figure retrieved by [38].}
\end{figure}

\textbf{Graphlets}. There are usually many components within a photo. Among these components, a few spatially neighboring ones and their interactions capture photo local aesthetics. Since a graph is a
powerful tool to describe the relationships between objects, [39] used graphs to model the spatial interactions between image components. Their technique is to segment a photo into a
set of atomic regions using unsupervised fuzzy clustering. Based on this, they extract graphlets Graphlet is a small-sized connected graph defined as: G = (V, E) where V is a set of vertices representing locally distributed atomic regions and E is a set of edges, each of which connects pairwise spatially adjacent atomic regions. Then, the graphlets are projected onto a
manifold and the authors subsequently proposed an embedding algorithm.

\textbf{Pyramid of Histograms of Orientation Gradients}. [26] proposed a model based on PHOG following a pyramid approach to calculate the HOG. HOG values are calculated based on the maximum gradient magnitudes in the color channels. Based on the new gradient image, we estimate the high self-similarity, moderate complexity and low anisotropy.

\textbf{Symmetry}. [30] measured the symmetry of an image based on the difference of the Histogram of Oriented Gradients (HOG) [7] between the first half of the image and its opposite right half.

\subsection{Content}

\textbf{Faces}. Recognition of active faces inside an image is still an open research field; the researchers of Computational Aesthetics borrow the advancements of this field in order to count the number of frontal faces, the relative size of the biggest face as well as the shadow area in faces. [20] The implementation used is usually the state of art algorithm proposed by [34].

\textbf{Skin}. Along with faces, skin color and recognition is important in order to identify a photograph containing humans. We estimate the number of skin pixels and the relative amount of skin with respect to the size of faces. The key concept is to detect the color space of pink that corresponds to human skin [19].

\section{Benchmark Datasets}
\label{S:3}

The above features are extracted usually from ground truth aesthetic images found in various dataset created for this task. Researchers test their techniques with the above benchmark datasets: DPChallenge, Photo.net, Flickr, Terragalleria, ALIPR and AVA. [14] 

\textbf{DPChallenge}\footnote{dpchallenge.com} allows users to participate and contest in theme-based photography on diverse themes such as life and death, portraits, animals, geology, street photography. Peer rating on overall quality, on a 1- 10 scale, determines the contest winners. (16,509 images) 

\textbf{Photo.Net}\footnote{photo.net} is a platform for photography enthusiasts to share and have their pictures peer rated on a 1 - 7 scale of aesthetics. The photography community also provides discussion forums, reviews on photos and photography products, and galleries for members and casual surfers. (20,278 images)  

\textbf{Flickr}\footnote{flickr.com} is one of the most popular online photo-sharing sites in the world. According to Flickr, the interestingness of a picture is dynamic and depends on a plethora of criteria including its photographer, who marks it as a favorite, comments, and tags given by the community. 

\textbf{Terragalleria}\footnote{terragalleria.com} displays travel photography of Quang-Tuan Luong (a scientist and a photographer), and is one of the finest resources for U.S. national park photography on the Web. All photographs here have been taken by one person (unlike other benchmarks), but multiple users have rated them on overall quality on a 1 - 10 scale. 

\textbf{ALIPR}\footnote{wang.ist.psu.edu/IMAGE}  is a Web-based image search and tagging system that also allows users to rate photographs along 10 different emotional categories such as surprising, amusing and adorable. 

\textbf{AVA}\footnote{lucamarchesotti.com} is the biggest aesthetics dataset which contains a rich variety of metadata including a large number of aesthetic scores for each image, semantic labels for over 60 categories as well as labels related to photographic style. (250,000 images) 

\section{Machine Learning}
\label{S:4}

Combining the above ground truth aesthetic datasets with the features extracted of their images, researchers employ machine learning techniques in order to predict the aesthetic value of an image. In this process, both classification and regression is performed for different desired results. With classification we can distinguish between an aesthetic photo or not. But this measure tends to be kind of absolute in such a subjective task. However, aesthetics is a more abstract
quality and quantifying it requires a relatively larger numeric scale. Thus, regression techniques provides us with a scale of aesthetic score for each photo, so we are able to compare why a photo is considered of better quality from another. SVM, SVR and classification and regression trees (CART) are the most usual methods used for classification and regression. In addition, forms of unsupervised learning like the K-means clustering are used for visual vocabulary generation and graph-based region segmentation,

The state of the art achieved on the above datasets varies according with the implementation and the training dataset size, making it difficult to compare the accuracy results of the proposed techniques. The best reported accuracies in most cases is around 80\% [8,35,39].

\section{Discussion}
\label{S:5}
The field of Computational Aesthetics seems promising considering the growing interest from the research community. Researchers try to model the aesthetic quality of photographs into low and high-level features, drawing inspiration from art theory, psychology and marketing. There are many quite imaginative applications of computational aesthetics such as coral reef evaluation image aesthetics [10], optimal text placement inside an image [18], better route suggestion in a city based on surroundings [25] or personality prediction [29] Some underexplored areas proposed by this review are: personalization feedback in predictions, non-photographic images, Black and White artistic consideration, and EXIF camera metadata..

\section{References}
\label{S:6}







\end{document}